\documentclass[11pt]{article}

\usepackage[final]{acl}

\usepackage{times}
\usepackage{latexsym}

\usepackage[T1]{fontenc}
\usepackage[utf8]{inputenc}

\usepackage{microtype}

\usepackage{inconsolata}

\usepackage{graphicx}

\usepackage{amsmath,amssymb}
\usepackage{enumitem}
\usepackage{booktabs}
\usepackage{url}
\usepackage{pifont}
\usepackage{makecell}

\newcommand{\cmark}{\ding{51}}% ✔
\newcommand{\xmark}{\ding{55}}% ✘

\graphicspath{{figures/}}

\title{REFLEX with Jev for Efficient Selective Control in LLM Agents}

\author{Tiantong Wu \\
  Nanyang Technological University\\Singapore \\
  \texttt{tiantong.wu@ntu.edu.sg} \\\And
  Wei Yang Bryan Lim \\
  Nanyang Technological University\\Singapore \\
  \texttt{bryan.limwy@ntu.edu.sg} \\}

\begin{document}

\maketitle

\begin{abstract}
LLM agents often use generative models for bounded decisions, raising the question of when these decisions can be handled more efficiently without reducing task success. We study \textsc{REFLEX}, an agent architecture that uses Jev as a fast, typed decision layer and calls a strong LLM when confidence is low, or generation is required. On a frozen 100-task benchmark, \textsc{REFLEX} achieves 95\% success with 72.7\% fewer strong-model calls than a strong-only agent, with reductions persisting across three fallback families. Controlled interventions show that reliability depends on action-set size and near-valid alternatives near authorization boundaries. External BFCL and $\tau$-style evaluations reveal limited advantages over a cheap generative cascade when ordinary routing is already highly accurate. These findings identify when selective control with Jev can reduce computation and where its benefits are limited.
\end{abstract}

\section{Introduction}
Tool-using language agents increasingly combine reasoning, natural-language generation, and actions through application programming interfaces (APIs). A common design uses a generative large language model (LLM) at nearly every step to inspect the state, select a tool, fill its arguments, decide whether to continue, and produce a response. This design is flexible, but it assigns different kinds of computation to the same model. Many steps require only a choice from a fixed set, suggesting an opportunity to reserve expensive generative reasoning for states that need it.

Existing approaches address related questions at different levels. Multi-LLM routers choose which model should answer an entire query \citep{chen2023frugalgpt,ong2024routellm,hu2024routerbench,li2026llmrouterbench}. Tool-use benchmarks primarily evaluate whether a model selects and executes functions correctly \citep{patil2025bfcl,wang2025dialogtool}. Our question concerns individual decisions within an agent's trajectory. Can a specialized decision model handle bounded choices while preserving task success, and under what conditions does it offer an advantage over a cheap generative cascade?

We study \textsc{REFLEX}, an agent architecture that uses Jev as its decision layer. Jev is a recently released ``System One'' model designed to make typed probabilistic decisions rather than generate strings \citep{typesafe2026jev}. In \textsc{REFLEX}, Jev returns a bounded choice and a confidence score. Choices with high confidence execute directly, while states with low confidence or requiring free-form generation are passed to the strong LLM. This separates bounded control from generative reasoning within an agent trajectory. Jev may select a tool or continue at the current step, while a strong LLM handles the next step if needed.

On a frozen 100-task controlled benchmark, \textsc{REFLEX} achieves 95\% success with 72.7\% fewer strong-model calls than a strong-only agent. Across Qwen3.8-Max, Kimi K3, and DeepSeek-V4-Pro fallbacks, it reduces strong-model calls by 66\% to 72\% while keeping success within a 2-point non-inferiority margin. These results show that selective control with Jev can improve the tradeoff between task success and computation across different strong fallback models.

Further experiments clarify which decisions remain difficult. On external BFCL tasks, function selection is already 98.4\% accurate, but deciding whether to call any function is only 52.0\% accurate. The harder question is therefore often whether the agent has enough authorization or information to act. Controlled interventions show that larger action sets and structurally near-valid alternatives reduce accuracy near authorization boundaries. Changing the type of alternative also shifts errors between irreversible commitment and deferral without a significant change in overall accuracy. These findings show why evaluating only whether the correct tool was selected can miss important differences in control reliability.

The external multi-turn evaluation also establishes a limit to the efficiency gains. On $\tau$-style tasks, \textsc{REFLEX} reduces cost by a factor of 3.7 relative to a strong-only agent, with a statistically unresolved difference in success. However, a cheap LLM cascade with self-escalation remains competitive. Together with the BFCL results, this suggests that a specialized decision layer has limited room to improve performance when ordinary routing is already highly accurate.

We make four contributions:
\begin{itemize}[leftmargin=*,nosep]
    \item We introduce \textsc{REFLEX}, an agent architecture that uses Jev for bounded decisions and calls a strong LLM when confidence is low, or generation is required.
    \item We show substantial reductions in strong-model calls on a frozen controlled benchmark across three fallback families, with success within a 2-point non-inferiority margin.
    \item We identify the conditions that limit the reliability of the decision layer. Larger action sets and near-valid alternatives reduce accuracy on decisions that depend on authorization, while the type of alternative causally shifts errors between unwarranted commitment and deferral.
    \item We establish the limits of the approach through external BFCL and $\tau$-style benchmark evaluations, showing that a cheap generative cascade remains competitive when ordinary routing is already highly accurate.
\end{itemize}

These findings show that the value of a specialized decision layer depends on the source of agent difficulty. Using Jev to separate bounded control from generation can reduce computation, but its advantage depends on whether the task contains control decisions that benefit from this separation.

\section{Related Work}
\paragraph{Cost-aware LLM routing.}
Cascades and routers use models with different costs and capabilities. FrugalGPT combines prompt adaptation, approximation, and cascades to reduce inference cost \citep{chen2023frugalgpt}. RouteLLM learns preference-based routers between strong and weak models \citep{ong2024routellm}. RouterBench and LLMRouterBench provide systematic evaluations of this setting. They show that models can complement one another, but consistently outperforming simple routing baselines remains difficult \citep{hu2024routerbench,li2026llmrouterbench}. These methods primarily choose a model for each query. \textsc{REFLEX} separates bounded control from generative reasoning within a single agent trajectory, allowing the two forms of inference to alternate during an episode.

\paragraph{Tool-using agents and evaluation.}
BFCL evaluates function selection, relevance detection, and, increasingly, multi-turn use of functions by agents \citep{patil2025bfcl}. DialogTool highlights the gap between single-turn tool selection and multi-turn interactions that maintain state \citep{wang2025dialogtool}. $\tau$-bench and $\tau^2$-bench evaluate conversational agents using domain policies and stateful tools. $\tau^2$ adds dual-control environments in which both the user and the agent modify shared state \citep{yao2024taubench,barres2025tau2}. We use these benchmarks to examine specific sources of difficulty. BFCL tests whether ordinary tool selection limits performance, while $\tau$ tests whether selective control improves complete trajectories.

\paragraph{Confidence and selective action.}
Selective prediction uses confidence to balance coverage and risk. Recent work examines confidence in tool-using agents, including model-internal confidence estimators \citep{subramani2025mice} and calibration failures that depend on the tool \citep{xuan2026confidence}. Other work provides causal evidence that language models use confidence to decide when to abstain \citep{zhou2026confidence}. \textsc{REFLEX} uses confidence to guide control. Bounded decisions with high confidence are executed directly, while states with low confidence are delegated to the fallback. We therefore report the relationship between risk and coverage in addition to calibration.

\section{\textsc{REFLEX}}
\label{sec:method}

\begin{figure*}[t]
\centering
\includegraphics[width=\textwidth]{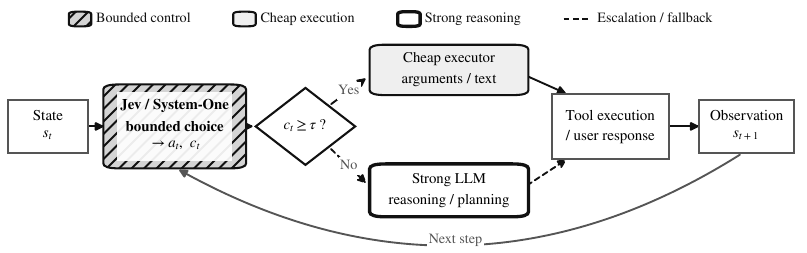}
\caption{\textsc{REFLEX} separates bounded agent control from open-ended generation. The decision layer reduces strong-model calls at states that can be handled as typed decisions, while retaining the strong LLM as a fallback.}
\label{fig:architecture}
\end{figure*}

\subsection{Agent control as a selective decision problem}
At agent step $t$, let $s_t$ denote the observable state and $\mathcal{A}_t$ a typed set of allowed bounded decisions, such as tools or control actions. A decision model $J$ returns
\begin{equation}
    \left(\hat a_t,c_t\right)=J\left(s_t,\mathcal{A}_t\right),
\end{equation}
where $c_t\in[0,1]$ is the confidence score. \textsc{REFLEX} applies
\begin{equation}
\pi(s_t)=
\begin{cases}
\hat a_t, & c_t\ge\tau\ \wedge\ \text{executable},\\
L_{\mathrm{strong}}(s_t), & \text{otherwise}.
\end{cases}
\label{eq:policy}
\end{equation}
The strong model handles open-ended reasoning, arguments that cannot be obtained deterministically, and user-facing text. In the realistic multi-turn evaluation, we place a cheap generative executor between Jev and the strong model because a typed choice model cannot generate text or arbitrary arguments.

The two outcomes of the gate serve different purposes. A decision with high confidence can remove a strong-model call, while a decision with low confidence passes control to the fallback without forcing an action. REFLEX can therefore retain the fallback's capabilities while changing which states require it. This motivates the cross-family experiment in \S~\ref{sec:results}. We test whether the decision model identifies a stable subset of states that it can handle autonomously without introducing new failures, rather than whether it can solve every task.

\subsection{Decision heads and execution semantics}
We distinguish three roles that are often combined in a single agent. \emph{Control} selects among bounded alternatives, such as which tool to call or whether the current state warrants action. \emph{Argument synthesis} supplies structured values for a selected action when those values are not already available in the state. \emph{Generation/reasoning} produces free-form text or resolves open-ended policy questions. REFLEX targets control. In REFLEX-Sim, many arguments are deterministic functions of the observable state. In the external multi-turn setting, a cheap generative executor supplies arguments and text after the control decision, while the strong model remains the final fallback.

This separation guides evaluation. We record tool-selection errors, argument errors, premature termination, and policy errors separately. We also distinguish \emph{invalid but recoverable} decisions from unsafe actions or actions that cause terminal failure. This allows us to distinguish a harmless extra retrieval from an irreversible write.

\paragraph{Compute metric.}
We measure the reduction in strong-model calls as
\begin{equation}
\mathrm{GMR}=1-\frac{N_{\mathrm{strong}}^{\textsc{REFLEX}}}{N_{\mathrm{strong}}^{\mathrm{strong-only}}}.
\end{equation}
We also report cost per episode, cost per successful episode, and strong calls per successful task. These measures help avoid rewarding methods that appear cheap because they fail early.

\paragraph{Selective reliability.}
For decisions executed autonomously at threshold $\tau$,
\begin{align}
\mathrm{Risk}(\tau) &= P\left(\text{invalid}\mid c\ge\tau\right),\\
\mathrm{Coverage}(\tau) &= P\left(c\ge\tau\right).
\end{align}
We summarize this tradeoff using the area under the risk-coverage curve (AURC) and compare it with random escalation at matched coverage.

\paragraph{Confidence gating and self-escalation.}
A cheap generative model can also decide when to call a stronger model. We include this approach as baseline B3. REFLEX is useful only if its typed decision signal identifies a subset that it can handle autonomously more efficiently or reliably than a cheap LLM cascade. We therefore treat B3 as a central baseline without assuming that a specialized decision layer will outperform it.

\section{Experimental Setup}
\label{sec:setup}

\paragraph{Models.}
The decision layer uses the version-pinned \texttt{jev-1.13.0} API. Qwen3.8-Flash and Qwen3.8-Max serve as the primary small and strong generative models. For cross-family validation, we replace the strong fallback with Kimi K3 or DeepSeek-V4-Pro while keeping Jev, the threshold, tools, and evaluator fixed. Kimi and DeepSeek use the same fixed high reasoning effort in the strong-only and REFLEX conditions.

\paragraph{REFLEX-Sim.}
We construct a deterministic tool environment that covers direct actions, semantic tool selection, short workflows, policy-governed actions, and ambiguous tasks requiring substantial reasoning (C0 to C4). We score tasks using terminal-state assertions and policy constraints rather than exact matches to a reference trajectory. After evaluator audits, we freeze the final version as \textsc{Reflex-Sim} v1.2.0. The test set contains 100 held-out tasks, and all reported main runs have zero harness errors.

We compare B0, a strong-only agent, B1, a Flash-only agent, B3, a Flash cascade with self-escalation, and R1, \textsc{REFLEX} with $\tau\in\{0.5,0.7,0.8,0.9,0.95\}$. The primary operating point is $\tau=0.5$. Later experiments keep this value fixed without repeating the threshold sweep.

\paragraph{Audit and freeze protocol.}
The controlled benchmark underwent two evaluator audits before the final v1.2.0 freeze. Success is determined by terminal state and policy constraints rather than agreement with a single gold action sequence, since valid agents may use shorter trajectories than the authored path. All frozen runs retain raw provider payloads, model and version manifests, task hashes, and exact tool traces. Offline grading corrections replay logged actions without regenerating model outputs. The main paper reports only the final frozen scores.

\paragraph{Inference and multiplicity.}
Unless otherwise stated, confidence intervals use paired cluster bootstrap resampling over the original task or scenario family. All perturbations derived from the same task or family remain within the resampled cluster. Final analyses use 10000 resamples, and exploratory studies use at least 2000. For paired binary episode success, we also use McNemar tests. The RF-5C factorial uses preregistered marginal cardinality, ambiguity, and interaction contrasts, avoiding selection of effects after inspecting the results. Paired uncertainty estimates use task-level bootstrap resampling.

\paragraph{Decision-complexity interventions.}
To examine the control layer, we vary action-set size $K$ and structural ambiguity. For action $a$ in state $s$, $d(a\mid s)$ counts the atomic state or policy facts that must change for $a$ to become correct. We compute this distance from declared conditions that must all hold for an action to be correct. Undeclared actions are classified as \textsc{far}. This defines levels of ambiguity independently of model predictions and embedding similarity. RF-5C crosses $K\in\{10,25,50\}$ with four ambiguity levels across 60 scenario families and two realizations per cell, yielding 1,440 decisions. We also run a matched intervention with 240 decisions. It replaces a one-change competitor that seeks information with one that makes an irreversible commitment, while holding the gold action, distance, competitor count, $K$, and semantic similarity fixed.

\paragraph{External evaluations.}
On BFCL \citep{patil2025bfcl}, we evaluate externally authored function-selection and function-relevance decisions. We expand $K$ using additional functions verified to be distant alternatives. In a $\tau$-style held-out multi-turn validation \citep{barres2025tau2}, we run 300 episodes comprising 60 paired tasks across five arms with a user simulator. We treat this as external validation rather than a leaderboard submission and use native terminal-state reward as the primary outcome. We audit step-level errors in control selection and argument generation separately.

\paragraph{Pricing and latency accounting.}
For every API call, we log input and output tokens, model identifiers, decoding parameters, and latency. Costs use dated price snapshots for each provider and region and are recomputed from raw token records. We report agent-side cost for the $\tau$ comparison because all arms share the same user simulator. Simulator cost is treated as benchmark overhead rather than a controller-specific difference. REFLEX-Sim also reports strong calls per successful task to account for apparent savings caused by early failure.

\section{Results}
\label{sec:results}

\subsection{REFLEX improves the tradeoff between success and computation}
\begin{figure}[t]
\centering
\includegraphics[width=\columnwidth]{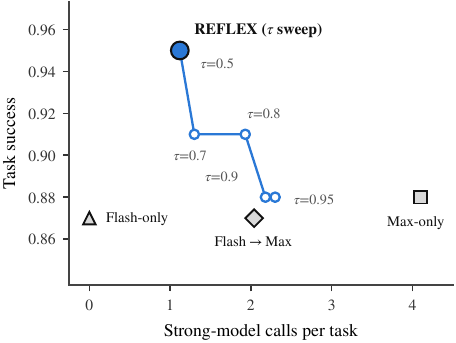}
\caption{Success and strong-model calls on the frozen 100-task REFLEX-Sim test set. R1 matches or exceeds B0's success rate at every threshold while using fewer strong-model calls. The largest gain occurs at $\tau=0.5$.}
\label{fig:frontier}
\end{figure}

Figure~\ref{fig:frontier} and Table~\ref{tab:main} show the corrected frozen results, with costs and additional efficiency measures in Table~\ref{tab:main_full}. B0 succeeds on 88\% of tasks with 4.10 strong calls per task. R1 at $\tau=0.5$ achieves 95\% success with 1.12 calls per task, a 72.7\% reduction. Higher thresholds reduce the success gain but still use fewer strong calls. At $\tau=0.9$ and $0.95$, R1 matches B0's 88\% success rate with 2.18 and 2.30 strong calls per task, respectively, corresponding to reductions of 46.8\% and 43.9\%. The observed success and call-count estimates therefore show a Pareto improvement over B0 at every tested threshold, without higher success at every threshold.

\begin{table}[t]\centering\small
\begin{tabular}{lccc}
\toprule
Condition & Success & Strong calls/task & GMR \\
\midrule
B0 Max-only & 0.88 & 4.10 & -- \\
B1 Flash-only & 0.87 & 0.00 & 1.000 \\
B3 Flash$\to$Max & 0.87 & 2.04 & 0.502 \\
R1 $\tau$=0.5 & \textbf{0.95} & 1.12 & 0.727 \\
R1 $\tau$=0.7 & 0.91 & 1.30 & 0.683 \\
R1 $\tau$=0.8 & 0.91 & 1.93 & 0.529 \\
R1 $\tau$=0.9 & 0.88 & 2.18 & 0.468 \\
R1 $\tau$=0.95 & 0.88 & 2.30 & 0.439 \\
\bottomrule
\end{tabular}
\caption{\textbf{Quality--compute frontier on frozen \textsc{reflex-sim} v1.2.0}
(100 held-out tasks). GMR is the reduction in strong-model calls relative to
B0. Bold marks the preferred frozen operating point ($\tau$=0.5), not a
best-in-column.}
\label{tab:main}
\end{table}

The selective-control metrics in Table~\ref{tab:selective} show that confidence is useful but imperfect. At $\tau=0.5$, all 65 episodes without escalation succeed. Of the 251 autonomous decisions on trajectories reached at that threshold, 250 are valid, with one harmless, recoverable exception. The reported AURC is 0.019, compared with 0.145 for random escalation at matched coverage, approximately 7.6$\times$ lower using the displayed values. Calibration is moderate, with a Brier score of 0.103 and ECE of 0.129. The main benefit is therefore the ability to distinguish decisions suitable for autonomous execution, despite imperfect calibration.

The paired outcomes in Table~\ref{tab:paired} help explain the gain. R1 and B0 both succeed on 87 tasks. R1 succeeds on 8 tasks that B0 fails and fails on 1 task that B0 succeeds on, while both fail on 4 tasks (exact McNemar $p=0.0391$).

Table~\ref{tab:replacement} breaks down the strong-model decisions that are replaced. Replacement rates are much higher for tool selection (99\%) and completion (77\%) than for clarification (28\%), with retrieval between them at 56\%. This suggests that routine bounded control is easier to delegate than clarification and judgments that depend on authorization.

\begin{table}[t]\centering\small
\begin{tabular}{lcc}
\toprule
Decision function & $n$ (B0 strong calls) & Replaced by Jev \\
\midrule
Tool selection & 116 & 99\% \\
Completion & 84 & 77\% \\
Retrieval decision & 160 & 56\% \\
Clarification & 18 & 28\% \\
\bottomrule
\end{tabular}
\caption{\textbf{Which strong-model decisions the control layer replaces}, on
frozen \textsc{reflex-sim}. Thin categories (policy reasoning $n$=7,
user-facing generation $n$=13, escalation $n$=12) are excluded from this
mechanistic claim; see Appendix for the full breakdown.}
\label{tab:replacement}
\end{table}

\subsection{Reliability depends on decision structure}
\begin{figure}[t]
\centering
\includegraphics[width=\columnwidth]{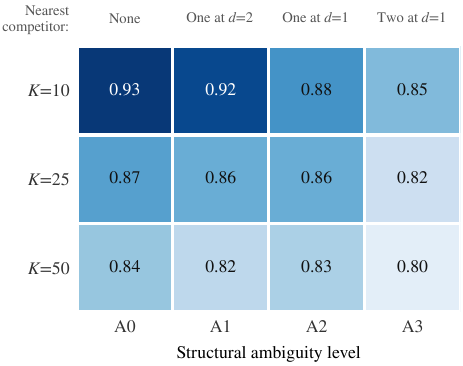}
\caption{RF-5C confirmatory factorial over 60 scenario families. The heatmap shows accuracy by action-set size and structural ambiguity. The cardinality and A0-to-A3 contrasts are significant, while their interaction is not (Table~\ref{tab:rf5c_full}).}
\label{fig:rf5c}
\end{figure}

Figure~\ref{fig:rf5c} shows the RF-5C factorial, and Table~\ref{tab:rf5c_full} reports the paired contrasts. Averaged over the four ambiguity levels, increasing $K$ from 10 to 50 reduces accuracy by 6.7 percentage points (95\% CI $[-11.3,-2.5]$, $p=0.0008$). Averaged over the three values of $K$, moving from A0 to A3 reduces accuracy by 5.6 percentage points (95\% CI $[-11.9,-0.3]$, $p=0.038$). The interaction is not statistically significant ($+3.3$ percentage points, 95\% CI $[-3.3,+10.8]$, $p=0.377$). Each of the 12 cells contains 120 decisions from 60 families with two surface realizations. Contrasts use family-level cell means, with paired cluster bootstrap resampling over families (Table~\ref{tab:rf5c_full}). The point estimate for increasing competitor count at fixed distance is larger in magnitude than that for reducing distance at fixed count ($-3.3$ versus $-1.1$ percentage points). However, both intervals include zero, with $p=0.099$ and $p=0.686$, respectively. These contrasts do not establish that competitor count has a stronger effect than distance.

These effects do not extend to all decision types. Table~\ref{tab:rf5c_errors} shows that 192 of the 208 errors occur on control decisions, comprising 117 risk errors and 75 clarification errors. The remaining 16 are completion errors. Retrieval, ordinary tool selection, and workflow decisions have zero errors across their combined 720 decisions. Table~\ref{tab:rf5c_full} reports 21 informative scenario families and 39 at ceiling. In its leave-risk-out sensitivity analysis, the cardinality estimate falls to $-2.7$ percentage points, with a 95\% interval of $[-6.7,+0.5]$ that includes zero. RF-5C, therefore, supports a narrower conclusion. Action-set size and near-valid alternatives matter when the central question is whether the agent has enough authorization or information to commit.

The construction audit in Table~\ref{tab:construct} also shows why embedding similarity is an inadequate measure of this form of ambiguity. Across 5,931 candidate actions, median similarity to the gold action is 0.310 for one-change competitors, 0.333 for two-change competitors, and 0.237 for far distractors. Moreover, 22.2\% of far distractors exceed the median similarity of the pooled near competitors, which is 0.322. These distributions show that similarity under the tested encoder does not consistently follow procedural distance, supporting a definition of ambiguity based on state and policy conditions.

\subsection{Competitor type changes the consequences of errors}
\begin{figure}[t]
\centering
\includegraphics[width=\columnwidth]{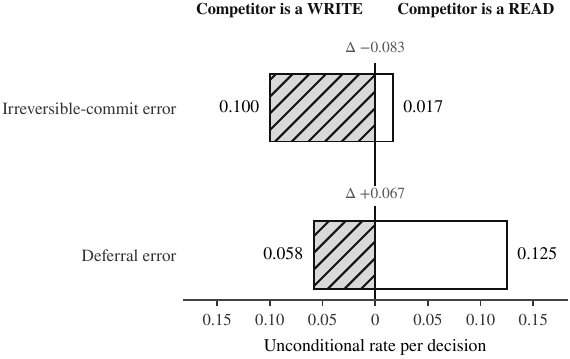}
\caption{Matched intervention on competitor type with $K$, gold action, counterfactual distance, competitor count, and semantic similarity held fixed. The intervention changes the \emph{type} of error without a significant change in overall accuracy.}
\label{fig:error_shape}
\end{figure}

Figure~\ref{fig:error_shape} and Table~\ref{tab:error_shape} report the matched intervention. Accuracy is 0.833 with a write competitor and 0.842 with a read competitor, a non-significant difference of $+0.008$ for read minus write (95\% CI $[-0.050,+0.067]$, $p=0.872$). Irreversible-commit errors occur on 10.0\% of decisions in the write arm and 1.7\% in the read arm ($\Delta=-0.083$, 95\% CI $[-0.150,-0.025]$, $p=0.0014$). Deferral errors occur on 5.8\% and 12.5\% of decisions, respectively ($\Delta=+0.067$, 95\% CI $[+0.025,+0.125]$, $p=0.0016$). All rates are unconditional per decision, and both differences use read minus write. Competitor type therefore changes the consequences of errors without a statistically resolved change in overall accuracy.

This intervention also revises the exploratory RF-5 observation that deferral was the characteristic failure. In that study, 219 of 366 errors were deferrals (59.8\%), and the deferral share rose from 0.36 to 0.72 as $K$ increased. In RF-5C, 24 of 208 errors are identified as deferrals (11.5\%, a lower bound), as reported in Table~\ref{tab:rf5c_errors}. The change across settings motivated retiring the general deferral claim. Table~\ref{tab:error_shape} reports no significant asymmetry between the shifts toward commitment and deferral ($\Delta=-0.017$, 95\% CI $[-0.067,+0.025]$, $p=0.534$). This does not establish equal effects or rule out a broader behavioral bias, but it shows that the available competitors influence the observed error pattern. Agents with similar exact-choice accuracy can therefore produce different downstream consequences.

\subsection{External benchmarks show where the benefits are limited}
\begin{figure*}[t]
\centering
\includegraphics[width=\textwidth]{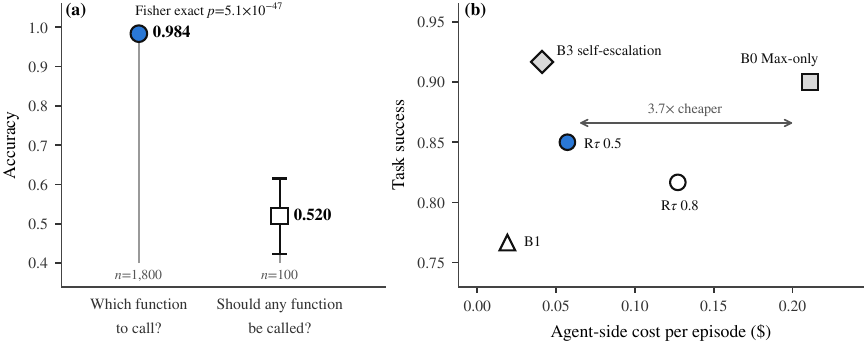}
\caption{External validation. Ordinary function selection on BFCL is nearly perfect, while deciding whether a function call is relevant remains much harder (left). On $\tau^2$-bench multi-turn tasks (right), REFLEX costs much less than Max-only, but a simple Flash cascade with self-escalation is competitive.}
\label{fig:external}
\end{figure*}

Figure~\ref{fig:external}(a) and Table~\ref{tab:bfcl} show a large gap between selecting a tool and deciding whether to act on BFCL.
With the same router and function schemas, choosing \emph{which} function to call achieves 98.4\% accuracy over 1,800 decisions. Deciding whether \emph{any} function should be called achieves only 52.0\% accuracy over 100 relevance cases (Fisher $p=5.1\times10^{-47}$). Increasing the candidate set from $K=2$ to $K=64$ changes routing accuracy from 99.0\% to 98.7\%. The paired difference is $-0.0033$ (95\% CI $[-0.010,0.000]$, $p=0.7202$), using a paired cluster bootstrap over 300 instances. None of the 29 errors selects an added distant distractor. Incorrect calls nevertheless have a mean confidence of 0.778, indicating that a $\tau=0.7$ gate would still allow incorrect calls. This external result supports the mechanism observed in RF-5C. The difficult decision is often whether to act, rather than which function to select.

Figure~\ref{fig:external}(b) and Table~\ref{tab:tau} show the corresponding system-level comparison on $\tau^2$-bench v1.0.1 across three domains, with 60 paired tasks evaluated under five conditions. REFLEX at $\tau=0.5$ reduces per-episode cost by a factor of 3.7 relative to B0 ($\$0.0572$ versus $\$0.2111$), with success of 0.850 versus 0.900. Table~\ref{tab:tau_pairwise} gives a paired success difference of $-0.050$ for REFLEX minus B0, with a 95\% interval of $[-0.150,+0.050]$ after reversing the reported contrast. B3 is cheaper than REFLEX ($\$0.0411$) and has numerically higher success (0.917 versus 0.850), but this paired difference is also unresolved ($+0.067$, 95\% CI $[-0.033,+0.167]$). Neither comparison establishes equal success. REFLEX at $\tau=0.8$ is dominated on the observed cost and success estimates by both B3 and REFLEX at $\tau=0.5$ and is not evaluated further.

The failure audit in Table~\ref{tab:tau2_failures} records 255 successful episodes and 45 failures. Of the failures, 41 are terminal database-state mismatches and four are communication failures. There are zero audited control-selection errors and zero argument-generation errors across all 300 episodes. Under these audit categories, the evaluated runs contain no observed routing or argument failures for the specialized decision layer to correct. The result therefore limits claims of a routing advantage on this evaluation, while showing a cost reduction relative to the strong-only baseline.

\begin{table}[t]\centering\small
\begin{tabular}{lccc}
\toprule
Condition & Success & \$/episode & \$/success \\
\midrule
B0 Max-only & 0.900 & 0.2111 & 0.235 \\
B1 Flash-only & 0.767 & 0.0191 & 0.025 \\
B3 self-escalation & \textbf{0.917} & \textbf{0.0411} & 0.045 \\
R$\tau$ $\tau$=0.5 & 0.850 & 0.0572 & 0.067 \\
R$\tau$ $\tau$=0.8 & 0.817 & 0.1272 & 0.156 \\
\bottomrule
\end{tabular}
\caption{Held-out 300-episode validation (60 paired tasks $\times$ 5
conditions). Only B3$-$B1 is significant among the reported pairwise
contrasts; other tested intervals span zero. Bold marks the best value in each
of the first two columns. No control-selection or argument-generation errors
occurred in any of the 300 episodes.}
\label{tab:tau}
\end{table}

\subsection{Strong-call reductions persist across fallback families}
\begin{figure*}[t]
\centering
\includegraphics[width=\textwidth]{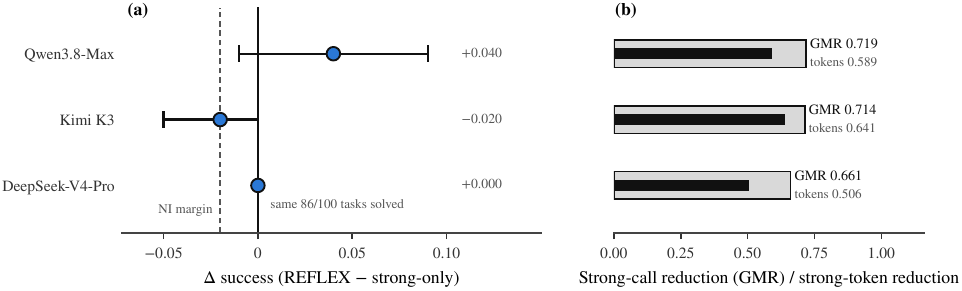}
\caption{Cross-family validation on the same frozen 100 tasks. Panel (a) shows paired success differences and 95\% bootstrap intervals. The dashed line marks the $-0.02$ margin, which Kimi's interval crosses. Panel (b) shows strong-call and strong-token reductions. Strong-model calls decrease by 66\% to 72\% across the three fallback families.}
\label{fig:e6}
\end{figure*}

Figure~\ref{fig:e6} and Tables~\ref{tab:e6} and~\ref{tab:e6_full} report the E6 comparison with Jev and $\tau=0.5$ fixed while the strong fallback changes. Strong-model calls decrease by 66.1\% to 71.9\%. The observed success differences for Qwen, Kimi, and DeepSeek are $+0.040$, $-0.020$, and $0.000$, respectively. Their 95\% bootstrap intervals are $[-0.010,+0.090]$, $[-0.050,0.000]$, and $[0.000,0.000]$. No family shows a statistically significant success decrease. However, the preregistered acceptance rule checked only whether the point estimate satisfied $\Delta\geq-0.02$. It was not a formal non-inferiority test. Under the confidence-bound rule requiring the lower 95\% bound to exceed $-0.02$, Qwen and DeepSeek meet the criterion, while Kimi does not (Table~\ref{tab:e6_mechanism}). Kimi's strong-only arm achieves 100\% success and REFLEX achieves 98\%, so its call reduction accompanies an observed two-task loss.

Figure~\ref{fig:e6}(b) and Table~\ref{tab:e6_full} also show strong-token reductions of 50.6\% to 64.1\%, indicating that call savings are accompanied by substantial token savings. E6 and the main threshold sweep are separate executions, and each is compared with its own baseline. Table~\ref{tab:run_to_run} documents their differences. Qwen strong-only success is 0.88 in the main run and 0.91 in E6, with six tasks lost and nine gained. Its strong-call rate changes from 4.10 to 3.95. REFLEX succeeds on 95 tasks in both runs, but four tasks are lost and four gained, while the autonomous subset changes from 65 to 66 tasks. The configurations differ only in the price table and Jev retry settings, neither of which affects B0. The B0 variation therefore occurs despite the same tasks, prompts, and temperature-zero inference condition. This provider variation limits the interpretation of small differences between separate runs.

\begin{table*}[t]\centering\small
\begin{tabular}{lcccc}
\toprule
Fallback family & Strong-only & REFLEX & $\Delta$ & GMR \\
\midrule
Qwen3.8-Max & 0.91 & 0.95 & $+0.04$ & 0.719 \\
Kimi K3 & 1.00 & 0.98 & $-0.02$ & 0.714 \\
DeepSeek-V4-Pro & 0.86 & 0.86 & $+0.00$ & 0.661 \\
\bottomrule
\end{tabular}
\caption{Cross-family robustness. Within each family the identical strong-model configuration (including reasoning effort) is used for both the
strong-only baseline and REFLEX. All individual success differences are
non-significant. The predeclared criterion compared the \emph{point estimate}
of $\Delta$ to a $-$0.02 margin; a formal non-inferiority test, which requires
the lower confidence bound to clear that margin, passes for Qwen and DeepSeek
but not for Kimi (Table~\ref{tab:e6_mechanism}). Kimi strong-only is at
ceiling, so that row cannot test quality improvement at all. Baseline success
here is not comparable to Table~\ref{tab:main}: both are the same condition
executed twice (Table~\ref{tab:run_to_run}).}
\label{tab:e6}
\end{table*}

Table~\ref{tab:subset} shows a task-level pattern that aggregate success does not reveal. For every fallback family, the same 66 of 100 tasks are handled entirely by Jev, and all 66 succeed. Every observed REFLEX failure in E6 occurs within the 34-task subset requiring escalation. For DeepSeek, strong-only and REFLEX succeed on exactly the same 86 tasks. REFLEX thus preserves the set of tasks solved by the fallback while removing two-thirds of its calls. The trace audit in Table~\ref{tab:e6_mechanism} further examines the failures and decisions within escalated episodes.

\subsection{What does the gate identify?}
\label{sec:gate_subset}

The stable subset in Table~\ref{tab:subset} provides information beyond aggregate success differences. The gate does not adapt to the identity or capability of the downstream strong model. With the decision layer and threshold fixed, the same autonomous subset is observed in all three E6 runs, and all 66 tasks succeed in each run. These rows describe one shared task subset, not three independent samples. This E6 subset is distinct from the 65 no-escalation episodes in the main threshold run reported in Table~\ref{tab:selective}. Differences in model capability appear on the remaining 34 tasks. Under strong-only execution, Kimi solves all of them, Qwen solves 76.5\%, and DeepSeek solves 58.8\%. REFLEX, therefore, filters computation according to the task rather than learning to approximate a particular fallback.

\begin{table*}[t]\centering\small
\begin{tabular}{lcccc}
\toprule
Fallback & Autonomous $n$ & Autonomous success & Escalated $n$
& Strong-only on escalated \\
\midrule
Qwen3.8-Max & 66 & 1.000 & 34 & 0.765 \\
Kimi K3 & 66 & 1.000 & 34 & 1.000 \\
DeepSeek-V4-Pro & 66 & 1.000 & 34 & 0.588 \\
\bottomrule
\end{tabular}
\caption{The fixed gate selects the same 66-task autonomous subset regardless of fallback family, and all observed E6 failures occur in the escalated subset. The three rows describe the same task subset under one frozen gate, not three independent samples.}
\label{tab:subset}
\end{table*}

GMR and autonomous task coverage use different denominators, so their numerical difference alone does not locate the savings. The trace analysis in Table~\ref{tab:e6_mechanism} provides direct evidence. Within episodes that escalate at least once, Jev still handles 26.2\% of steps for Qwen, 26.8\% for Kimi, and 21.3\% for DeepSeek without a strong-model call. The savings therefore extend beyond the 66 fully autonomous tasks. Across the three REFLEX arms, the audit identifies 21 failures, comprising five for Qwen, two for Kimi, and 14 for DeepSeek. None contains an incorrect autonomous decision. The audit attributes these observed failures to fallback handling after escalation rather than an incorrect decision executed by Jev.

\section{Discussion}
The experiments show that the value of specialized decision layers depends on the setting. First, bounded control can serve as a separate form of computation. On REFLEX-Sim, a stable 66\% subset is solved autonomously with zero failures across three distinct fallback families, while the strong model handles the remaining states. This substantially reduces computation without requiring the decision model to reproduce the fallback's full reasoning behavior.

The cross-family results show large call reductions without a statistically significant success decrease, but this does not establish non-inferiority for every family. Qwen and DeepSeek meet the lower-confidence-bound criterion, while Kimi does not. The repeated Qwen runs also show a three-percentage-point change in strong-only success despite the same inference condition (Table~\ref{tab:run_to_run}). This observed change is comparable in scale to the E6 success differences. It is evidence of run-to-run variability, not an estimated uncertainty band, and limits claims about small quality differences from a single execution.

Second, the advantage depends on the source of difficulty. BFCL shows that the tested decision model already selects the correct function almost perfectly, while deciding whether a call is warranted remains harder. The $\tau^2$-bench validation extends this finding to complete trajectories. When no tool-selection or argument errors are observed, a dedicated router has little room to improve performance, and a simple, cheap cascade with self-escalation is competitive. This agrees with broader LLM-routing evidence that more complex routers do not consistently outperform simple baselines under a common evaluation \citep{li2026llmrouterbench}.

Third, accuracy alone can hide differences in control errors. Near authorization boundaries, larger candidate sets and greater structural ambiguity reduce reliability. Changing the \emph{type} of one near-valid competitor also shifts errors between irreversible commitment and deferral without a significant change in accuracy. Evaluation of selective agents should therefore report action consequences and the types of risk, as well as exact-choice accuracy.

These results suggest three conditions for using specialized decision layers. A meaningful part of the trajectory should consist of bounded, repeated decisions. Confidence should distinguish states suitable for autonomous execution from those that require escalation. The downstream benchmark should also account for uncertainty in control decisions. Otherwise, the additional component may increase complexity without improving the cost-success trade-off.

The full claim register in Table~\ref{tab:claims} records the supported and retired claims, including the qualified cross-family and $\tau^2$-bench comparisons.

\paragraph{Why the negative external result matters.}
REFLEX-Sim alone would suggest broad benefits from the architecture. BFCL and $\tau^2$-bench show that the room for improvement depends on the task. This matters for both deployment and evaluation. A benchmark with almost no routing errors cannot establish the value of a new routing layer, and overall agent success does not reveal whether failures arise from control, arguments, or downstream reasoning. The negative external result narrows the claim and makes its limits clearer and more directly testable.

\paragraph{Implications for agent benchmarks.}
Across our experiments, ordinary tool selection often reaches ceiling performance, while decisions that depend on authorization remain difficult. Future agent benchmarks should therefore annotate both the correct action and why it is currently valid. Relevant information includes the required evidence, authorization state, reversibility, and whether seeking more information is still warranted. These annotations would support the evaluation of selective risk without requiring one reference trajectory.

\section{Conclusion}
We presented \textsc{REFLEX}, which separates bounded control from open-ended reasoning in LLM agents. In the cross-family comparison, it reduces strong-model calls by roughly two-thirds across Qwen, Kimi, and DeepSeek fallbacks, with observed success differences ranging from a two-percentage-point loss to a four-percentage-point gain. No success decrease is statistically significant, but only Qwen and DeepSeek meet the confidence-bound non-inferiority criterion in these runs. Controlled interventions show that reliability is mainly limited by commitment decisions that depend on authorization, rather than ordinary tool selection. BFCL and $\tau^2$-bench provide a complementary negative result. Cheap-model cascades are competitive when routing is already nearly perfect. The value of a decision layer, therefore, depends on whether its form of inference fits the computation needed at each step.

\section*{Limitations}
We use one proprietary typed decision model, Jev, so the results do not establish that all non-generative decision models have the same selective-control behavior. E6 varies the strong fallback across three families but keeps the decision model fixed. Several controlled benchmarks also show ceiling effects for 2026 frontier models. Kimi reaches 100\% success on REFLEX-Sim, and ordinary BFCL function selection is nearly perfect. The effects of action-set size and ambiguity in RF-5C are concentrated in authority and risk families and should not be generalized to all agent decisions. The $\tau$ evaluation is a held-out validation with 300 episodes, rather than a full leaderboard evaluation with multiple trials. Its native reward can also miss some procedural differences. We therefore use it as external evidence about cost and where failures occur, without claiming benchmark superiority. Finally, reported API costs are dated snapshots and may change with provider pricing, caching, or deployment region.

\bibliography{custom}

\appendix

\setcounter{table}{0}
\renewcommand{\thetable}{S\arabic{table}}
\ifdefined\theHtable
\renewcommand{\theHtable}{supp.\arabic{table}}
\fi

\section{Hierarchical Routing Ablation}
\label{app:hierarchy}

We tested a hierarchical routing variant that first selects a tool family and then selects a tool within that family. This design reduces the number of alternatives considered at the second stage, but introduces a separate family-selection decision. The ablation examines whether this additional stage helps when the candidate set is large.

An initial smoke study suggested that hierarchical routing performed worse than flat routing. However, the implementation audit identified differences in the information shown to the compared systems (Table~\ref{tab:audit}). The family-selection stage received bare family names without their authored descriptions, while the router described only the original pilot tools rather than every candidate. In addition, a generative model prompt silently omitted candidates without a known specification. These issues weakened the hierarchical arm and affected the interpretation of the flat-routing comparison. The affected smoke results, therefore, do not support a clean comparison between routing structures.

The repair supplied the authored family descriptions, described all candidates, and retained candidates by identifier when a specification was unavailable. After these repairs, hierarchical accuracy improved by 5 to 10 percentage points relative to the broken implementation. This improvement measures the effect of repairing the implementation and does not establish an advantage over corrected flat routing. Conditional on selecting the correct family, within-family tool selection was perfect in the evaluated probes. The remaining hierarchical errors were concentrated at family boundaries, indicating that the first stage remained a source of failure.

Corrected flat routing was already near the ceiling on well-posed single-step probes, including those with large candidate sets. Splitting a decision into two stages, therefore, offered limited room for improvement while introducing a potentially ambiguous family assignment. An action may plausibly fit more than one semantic category, so reducing the number of tools considered does not necessarily make the full decision easier. We consequently treat hierarchy as a negative ablation rather than part of \textsc{REFLEX}. The result motivated RF-5C's controlled study of action-set size and structural ambiguity instead of a larger hierarchy sweep. The general claim that hierarchical routing helps at large candidate-set sizes is retired in Table~\ref{tab:claims}.

\section{Additional Evaluation Results}
\label{app:verification}

\subsection{Selective reliability and confidence}
\label{app:selective}

Table~\ref{tab:selective} reports operational metrics from each executed threshold run. These measurements describe the trajectories reached under that threshold, rather than counterfactual decisions pooled across runs. Decision risk is the fraction of autonomous decisions that are invalid. Episode-level MER instead measures the fraction of episodes containing at least one invalid autonomous decision. A recoverable invalid decision can therefore increase both measures without causing the episode to fail.

At $\tau=0.5$, 250 of 251 autonomous decisions are valid, giving a decision risk of 0.0040. The single invalid decision is recoverable, and all 65 episodes completed without escalation succeed. Increasing the threshold reduces decision coverage from 0.691 at $\tau=0.5$ to 0.439 at $\tau=0.95$. No unsafe autonomous decision is observed at any reported threshold. These are empirical outcomes on the evaluated runs, not guarantees of error-free autonomous execution.

Jev's AURC is 0.019, compared with 0.145 for random escalation at matched coverage. Its Brier score of 0.103 and ECE of 0.129 show that useful selective discrimination does not require perfect calibration. B3 does not emit a numeric confidence score, so an equivalent confidence-based risk-coverage curve is unavailable for that baseline.

\begin{table*}[t]\centering\small
\begin{tabular}{lcccccc}
\toprule
$\tau$ & Coverage & Decision risk & UnsafeRisk & NES & RecoveryRate
& Episode-level MER \\
\midrule
0.5 & 0.691 & 0.0040 & 0.0000 & 1.000 & 0.857 & 0.010 \\
0.7 & 0.629 & 0.0045 & 0.0000 & 1.000 & 0.800 & 0.010 \\
0.8 & 0.514 & 0.0049 & 0.0000 & 1.000 & 0.827 & 0.010 \\
0.9 & 0.468 & 0.0000 & 0.0000 & 1.000 & 0.769 & 0.000 \\
0.95 & 0.439 & 0.0000 & 0.0000 & 1.000 & 0.769 & 0.000 \\
\bottomrule
\end{tabular}
\caption{\textbf{Operational selective-control metrics}, each computed from its
own executed threshold run (no pooling of counterfactual replays). Decision
risk is the invalid rate among autonomous decisions; episode-level MER is the
fraction of episodes containing at least one invalid autonomous decision.
The two levels are reported separately and are not the same statistic: at
$\tau$=0.5, 250 of 251 autonomous decisions were valid
(99.6\%), and the single invalid decision was recoverable, while 65/65 no-escalation
episodes succeeded (100\%). Decision-level
validity is not perfect; episode-level no-escalation success is.
Jev AURC = 0.019; random matched-coverage AURC = 0.145; Brier = 0.103; ECE = 0.129. A comparable curve for
the B3 router is unavailable: it emits no numeric confidence.}
\label{tab:selective}
\end{table*}

\subsection{RF-5C design and statistical contrasts}
\label{app:rf5c_statistics}

RF-5C crosses three action-set sizes, $\mathcal{K}=\{10,25,50\}$, with four ambiguity levels, A0 to A3. Each of the 60 scenario families contributes two surface realizations to each cell, yielding 120 decisions per cell and 1,440 decisions overall. The ambiguity levels contain no near competitor, one two-change competitor, one one-change competitor, and two one-change competitors, respectively.

Let $m_f(K,A)$ be the mean correctness of the two surface realizations for family $f$ in cell $(K,A)$, and let $\mathcal{L}=\{\mathrm{A0},\mathrm{A1},\mathrm{A2},\mathrm{A3}\}$. With $F=60$, the reported cardinality and ambiguity contrasts are
\begin{align}
\Delta_K &= \frac{1}{4F}\sum_{f=1}^{F}\sum_{A\in\mathcal{L}}\left[m_f(50,A)-m_f(10,A)\right],\\
\Delta_A &= \frac{1}{3F}\sum_{f=1}^{F}\sum_{K\in\mathcal{K}}\left[m_f(K,\mathrm{A3})-m_f(K,\mathrm{A0})\right].
\end{align}
Thus, the ambiguity contrast averages over all three action-set sizes rather than conditioning on $K=25$. The cardinality contrast averages over all four ambiguity levels. Confidence intervals use 10,000 paired cluster bootstrap resamples over scenario families with seed 20260921. Each resampled family retains its complete set of cells. The fixed contrast weights are recorded in \texttt{rf5c/inference\_plan.json}.

Table~\ref{tab:rf5c_full} reports significant marginal decreases for cardinality and ambiguity, but not their interaction. The count-at-fixed-distance and distance-at-fixed-count estimates are also individually non-significant. Their different point estimates, therefore, do not establish which component has the larger effect. The leave-risk-out result is a sensitivity analysis rather than a preregistered primary contrast. Its interval includes zero, showing that the aggregate cardinality result depends on the decision families represented in the panel.

\begin{table*}[t]\centering\small
\begin{tabular}{lccc}
\toprule
Contrast & Estimate & 95\% CI & $p$ \\
\midrule
$K$ 10$\to$50 at mean ambiguity & $-0.067$ & $[-0.113, -0.025]$ & 0.0008 \\
A0$\to$A3 at $K$=25 & $-0.056$ & $[-0.119, -0.003]$ & 0.038 \\
$K \times$ A interaction & $+0.033$ & $[-0.033, +0.108]$ & 0.377 \\
\quad distance at fixed count & $-0.011$ & $[-0.061, +0.031]$ & 0.686 \\
\quad count at fixed distance & $-0.033$ & $[-0.081, +0.003]$ & 0.099 \\
\midrule
leave-risk-out ($K$ effect) & $-0.027$ & $[-0.067, +0.005]$ & -- \\
\bottomrule
\end{tabular}
\caption{\textbf{RF-5C factorial statistics.} 60 scenario families, of which 21
are informative and 39 sit at ceiling. 1{,}440 decisions in 12 cells of 120
(60 families $\times$ 2 surface realizations). The estimator is a paired
cluster bootstrap over scenario family, 10{,}000 resamples, seed 20260921:
the two surface realizations are first averaged into each family's cell mean,
then each contrast is a fixed weighted sum of those cell means, with weights
given in \texttt{rf5c/inference\_plan.json}. The $K$ contrast is the mean over
the four ambiguity levels of (K=50 $-$ K=10), i.e. $0.825-0.892=-0.067$ from
the cell means in Figure~\ref{fig:rf5c}; the ambiguity contrast is the mean
over the three $K$ levels of (A3 $-$ A0). The leave-risk-out row shows the
aggregate effect is carried by control/authority-sensitive families and must
not be generalized across all decision types (Table~\ref{tab:rf5c_errors});
Its $p$ is not reported because that refit is a sensitivity analysis, not a
pre-registered contrast.}
\label{tab:rf5c_full}
\end{table*}

\subsection{Matched intervention on competitor type}
\label{app:competitor_type}

The matched intervention contains 240 decisions at $K=25$. It changes whether the single one-change competitor is an information-seeking read or an irreversible write, while holding the gold action, competitor count, procedural distance, and semantic similarity fixed. Table~\ref{tab:error_shape} reports every difference as the read-arm rate minus the write-arm rate. Error rates are unconditional per decision, not conditional on making an error.

Accuracy is 0.833 in the write arm and 0.842 in the read arm. The difference is unresolved, with a 95\% interval of $[-0.050,+0.067]$. However, the read arm has fewer irreversible-commit errors and more deferral errors. The corresponding differences are $-0.083$ and $+0.067$, and both intervals exclude zero. Competitor type, therefore, changes the consequences of errors even when the overall accuracy difference is not statistically resolved. The asymmetry test has $p=0.534$, which does not establish that the two shifts are equal or that the model has no broader behavioral bias.

\begin{table*}[t]\centering\small
\begin{tabular}{lccccc}
\toprule
Outcome & Commit-shaped & Defer-shaped & $\Delta$ (defer$-$commit) & 95\% CI & $p$ \\
\midrule
Accuracy & 0.833 & 0.842 & $+0.008$ & $[-0.050, +0.067]$ & 0.872 \\
Irreversible-commit error & 0.100 & 0.017 & $-0.083$ & $[-0.150, -0.025]$ & 0.0014 \\
Deferral error & 0.058 & 0.125 & $+0.067$ & $[+0.025, +0.125]$ & 0.0016 \\
Shape-pull asymmetry & -- & -- & $-0.017$ & $[-0.067, +0.025]$ & 0.534 \\
\bottomrule
\end{tabular}
\caption{\textbf{Matched competitor-shape intervention}, 240 decisions at
$K$=25. Arms differ only in whether the single one-change competitor is a read
or a write; all rates are unconditional per decision.}
\label{tab:error_shape}
\end{table*}

\subsection{BFCL function selection and relevance}
\label{app:bfcl}

Table~\ref{tab:bfcl} separates function selection from the decision of whether a function applies. The cardinality evaluation contains 300 instances at each of six candidate-set sizes, for 1,800 decisions in total. Added functions are certified as topically distant and checked through a manual sample audit. The resulting comparison tests expansion with distant alternatives and do not establish robustness to arbitrary near-valid competitors.

The paired accuracy difference between $K=64$ and $K=2$ is $-0.0033$, with a 95\% interval of $[-0.010,0.000]$ and $p=0.7202$. Resampling is paired and clustered over the 300 instances. None of the 29 incorrect selections chooses an injected distant distractor. In contrast, accuracy on the separate 100-case relevance evaluation is 0.520, compared with 0.984 for function selection. Incorrect calls have a mean confidence of 0.778, indicating that a threshold of 0.7 cannot exclude all incorrect calls. The construction uses procedural checks because high embedding similarity alone cannot establish whether a function is a valid alternative.

\begin{table*}[t]\centering\small
\begin{tabular}{lcc}
\toprule
Stratum & Accuracy & $n$ \\
\midrule
\multicolumn{3}{l}{\emph{Routing cardinality}} \\
$K$=2 & 0.990 & 300 \\
$K$=4 & 0.990 & 300 \\
$K$=8 & 0.980 & 300 \\
$K$=16 & 0.973 & 300 \\
$K$=32 & 0.983 & 300 \\
$K$=64 & 0.987 & 300 \\
\midrule
\multicolumn{3}{l}{\emph{Relevance}} \\
Which function to call & 0.984 & 1{,}800 \\
Should any function be called & 0.520 & 100 \\
False-positive abstention & 0.010 & 100 \\
Mean confidence of wrong calls & 0.778 & -- \\
\midrule
\multicolumn{3}{l}{\emph{Distractor audit}} \\
Errors selecting an injected distant distractor & 0 / 29 & -- \\
\bottomrule
\end{tabular}
\caption{\textbf{BFCL.} Controlled fill was certified topically distant and
hand-audited on a random sample. No plausible-distractor regime is reported:
BFCL's native distractors reach 0.94 similarity to their gold, so similarity
alone cannot certify that an injected function is irrelevant.}
\label{tab:bfcl}
\end{table*}

\subsection{Paired comparisons on $\tau^2$-bench}
\label{app:tau2_pairwise}

The external multi-turn evaluation uses $\tau^2$-bench v1.0.1 across three domains. Sixty held-out tasks are evaluated under five conditions, yielding 300 episodes. Table~\ref{tab:tau_pairwise} uses paired cluster bootstrap resampling over tasks with 10,000 resamples. Pairing preserves the comparison between conditions on the same task.

Among the reported contrasts, only B3 minus B1 has an interval excluding zero. B3 exceeds REFLEX at $\tau=0.5$ by an observed 0.067, but the interval is $[-0.033,+0.167]$. B0 exceeds REFLEX by 0.050, with an interval of $[-0.050,+0.150]$. These unresolved differences do not establish equal performance. REFLEX's lower episode cost relative to B0 can therefore be reported separately from the uncertainty in the difference in their success. The failure categories underlying these outcomes are given in Table~\ref{tab:tau2_failures}.

\begin{table*}[t]\centering\small
\begin{tabular}{lccc}
\toprule
Contrast & $\Delta$ success & 95\% CI & Significant? \\
\midrule
B3 $-$ B1 & $+0.150$ & $[+0.033, +0.267]$ & yes \\
B3 $-$ R$\tau$ 0.5 & $+0.067$ & $[-0.033, +0.167]$ & no \\
B0 $-$ R$\tau$ 0.5 & $+0.050$ & $[-0.050, +0.150]$ & no \\
R$\tau$ 0.5 $-$ R$\tau$ 0.8 & $+0.033$ & $[-0.067, +0.133]$ & no \\
B3 $-$ B0 & $+0.017$ & $[-0.083, +0.117]$ & no \\
\midrule
\multicolumn{4}{l}{\footnotesize Audit: $E_{\text{control}}=0$,
$E_{\text{arguments}}=0$ across all 300 episodes.} \\
\bottomrule
\end{tabular}
\caption{Paired cluster bootstrap over the 60 held-out tasks, 10{,}000
resamples. Only B3$-$B1 excludes zero.}
\label{tab:tau_pairwise}
\end{table*}

\subsection{Cross-family calls, tokens, and costs}
\label{app:e6_full}

Table~\ref{tab:e6_full} reports the E6 results for each strong fallback. Within a family, the strong-only and REFLEX arms use the same strong-model configuration, including reasoning effort. Jev and the threshold remain fixed across families. Strong-call reductions are 71.9\% for Qwen, 71.4\% for Kimi, and 66.1\% for DeepSeek. Strong-token reductions are 58.9\%, 64.1\%, and 50.6\%, respectively.

Call reduction, token reduction, and monetary cost reduction are different quantities. A removed call can vary in length, and the remaining calls can have a different token distribution. Verified cost reduction is reported only for Qwen at 52.2\%. Missing Kimi and DeepSeek cost entries indicate that a verified frozen provider price is unavailable, not that cost reduction is zero. The success intervals and their implications for non-inferiority are examined further in Table~\ref{tab:e6_mechanism}.

\begin{table*}[t]\centering\small
\begin{tabular}{lccccccccc}
\toprule
Strong family & Strong-only & REFLEX & $\Delta$ & 95\% CI & \makecell{B0\\calls/task}
& \makecell{REFLEX\\calls/task} & GMR & Token red. & Cost red. \\
\midrule
Qwen3.8-Max & 0.91 & 0.95 & $+0.04$ & $[-0.01, +0.09]$ & 3.95 & 1.11 & 0.719 & 0.589 & 0.522 \\
Kimi K3 & 1.00 & 0.98 & $-0.02$ & $[-0.05, +0.00]$ & 3.53 & 1.01 & 0.714 & 0.641 & -- \\
DeepSeek-V4-Pro & 0.86 & 0.86 & $+0.00$ & $[+0.00, +0.00]$ & 4.04 & 1.37 & 0.661 & 0.506 & -- \\
\bottomrule
\end{tabular}
\caption{Full E6 results. Cost reduction is reported only where a verified
frozen provider price exists. \emph{Kimi strong-only is at ceiling (1.00), so
that row cannot test quality improvement.} The predeclared non-inferiority
criterion was applied to the point estimate of $\Delta$, not to the
confidence bound; Table~\ref{tab:e6_mechanism} gives the formal test.}
\label{tab:e6_full}
\end{table*}

\section{Implementation Audits and Verification}
\label{app:audits}

\subsection{Corrections before the final analyses}
\label{app:implementation_audit}

Table~\ref{tab:audit} records the issues found during development, their potential effects, and the evidence affected. The pilot evaluator initially required an exact ticket-reason string, which could penalize otherwise valid outcomes. Other pilot issues included unobservable eligibility conditions, omitted action histories, and argument templates that covered only the authored path. These issues were corrected through evaluator changes, task repairs, and more complete agent inputs. Offline grading corrections used recorded trajectories rather than regenerating model outputs.

The table also records the candidate-description and prompt-display errors relevant to the hierarchical ablation. Later checks addressed cache keys that omitted decoding parameters, Jev tokens priced at strong-model rates, and tie handling in the BFCL AUROC calculation. Entries marked as affecting no frozen evidence were found before the corresponding module was frozen. The floating-point tolerance repair applies to a point-estimate acceptance check and does not turn that check into a statistical non-inferiority test.

\begin{table*}[t]\centering\footnotesize
\begin{tabular}{p{0.10\textwidth}p{0.24\textwidth}p{0.20\textwidth}p{0.24\textwidth}p{0.11\textwidth}}
\toprule
Stage & Issue found & Potential bias & Resolution & Frozen evidence affected \\
\midrule
Pilot grading & Exact-string \texttt{create\_ticket} reason match & Inflated recovery counts & Evaluator relaxed; offline regrade & regraded to v1.2.0 \\
Pilot task design & C3 eligibility unobservable from state & Ill-posed decision & Observable fields added; tasks rebuilt & pre-v1.2.0 only \\
RF-5a & E1a: only pilot tools described to the router & Weakened added distractors; biased $K$ sweep flat & All candidates described; parity logged per call & RF-5a smoke only \\
RF-5a & E1b: family stage sent bare family names & Understated hierarchical arm & Authored descriptions wired in & RF-5a smoke only \\
RF-5a & E1c: LLM prompt silently dropped unknown candidates & Model graded on options it never saw & Never drop; list by id if unspecced & RF-5a Flash arm void \\
Pilot agents & Prompt omitted action history & Repeated-action loops & History added & pre-v1.2.0 only \\
Pilot agents & Argument templates covered gold-path tools only & Non-gold tools always failed & Templates extended; LLMs propose args & pre-v1.2.0 only \\
E6 & Provider cache key omitted decoding parameters & A reply from one reasoning effort could serve another & Decode params and \texttt{extra\_body} folded into the key & none \\
E5/E6 & Jev tokens priced at the strong-model rate & Overstated REFLEX cost & Tokens split by controller & none \\
BFCL & AUROC ignored ties & Sign-flip asymmetry in a tied score & Mid-ranks & none \\
E6 & Non-inferiority check used a literal $\geq -0.02$ & Criterion reported false on a float artefact & Tolerance added & none \\
\bottomrule
\end{tabular}
\caption{\textbf{Audit history.} Issues that materially affected
interpretation, and which frozen evidence each one touches. Items marked
``none'' were found before the affected module was frozen.}
\label{tab:audit}
\end{table*}

\subsection{Paired outcomes on REFLEX-Sim}
\label{app:paired_outcomes}

Table~\ref{tab:paired} gives the full paired outcome table for B0 and REFLEX at $\tau=0.5$ on the 100-task main test set. Both succeed on 87 tasks. REFLEX succeeds on eight tasks that B0 fails and fails on one task that B0 succeeds on. Both fail on four tasks. The margins are therefore 88 B0 successes and 95 REFLEX successes. The exact two-sided McNemar test uses the nine discordant pairs and gives $p=0.0391$. This pairing supports a net improvement in the observed main run while retaining the one observed regression.

\begin{table}[t]\centering\small
\begin{tabular}{lccc}
\toprule
& \multicolumn{2}{c}{\textsc{reflex} $\tau$=0.5} & \\
\cmidrule(lr){2-3}
$B_0$ Max-only & succeeds & fails & total \\
\midrule
succeeds & 87 & 1 & 88 \\
fails & 8 & 4 & 12 \\
\midrule
total & 95 & 5 & 100 \\
\bottomrule
\end{tabular}
\caption{\textbf{Paired task outcomes} on the frozen 100-task
\textsc{reflex-sim} test split. The gate recovers 8 tasks
$B_0$ fails and loses 1, so the marginal move from 0.88 to 0.95 is not a trade of one failure
set for another. McNemar exact $p$=0.039 on the 9 discordant pairs.}
\label{tab:paired}
\end{table}

\subsection{RF-5C error concentration and exploratory deferrals}
\label{app:rf5c_errors}

Table~\ref{tab:rf5c_errors} separates the 1,440 RF-5C decisions into six types, each containing 240 decisions. Risk contributes 117 errors and clarification contributes 75, so control decisions account for 192 of the 208 errors. The remaining 16 are completion errors. Retrieval, ordinary tool selection, and workflow have no errors across their combined 720 decisions. The full analysis still includes all 60 scenario families, although 39 remain at ceiling and 21 are informative for the observed errors.

The exploratory RF-5 study recorded 219 deferrals among 366 errors (59.8\%), with the share increasing from 0.36 to 0.72 as $K$ increased. RF-5C identifies 24 deferrals among 208 errors (11.5\%), reported as a lower bound. These are different study settings and do not form a matched estimate of a change in deferral rate. Their disagreement, together with the competitor-type intervention, limits the claim that deferral is a general failure tendency of the router.

\begin{table*}[t]\centering\small
\begin{tabular}{llccc}
\toprule
Decision type & Gold kind & Decisions & Errors & Accuracy \\
\midrule
risk & control & 240 & 117 & 0.512 \\
clarification & control & 240 & 75 & 0.688 \\
completion & terminate & 240 & 16 & 0.933 \\
retrieval & gather & 240 & 0 & 1.000 \\
tool selection & gather & 240 & 0 & 1.000 \\
workflow & commit & 240 & 0 & 1.000 \\
\midrule
all & & 1440 & 208 & 0.856 \\
\bottomrule
\end{tabular}
\caption{\textbf{RF-5C errors by decision type.} 192 of 208 errors (92\%) fall on
decisions whose gold action is a \emph{control} action (risk and
clarification); the rest are completion decisions. Retrieval, ordinary tool
selection and workflow are at a perfect ceiling across all 720 of their decisions, which is
why the effective corpus is 21 families rather than 60. Each of the 12
factorial cells contains 120 decisions (60 families $\times$ 2 surface
realizations); deferral accounts for 24 of the 208 errors here (11.5\%, a lower bound) against 59.8\% of 366 errors in the
exploratory study, which is the disagreement that retired C4.}
\label{tab:rf5c_errors}
\end{table*}

\subsection{Procedural distance and embedding similarity}
\label{app:construct}

Table~\ref{tab:construct} examines all 5,931 candidate actions from the 60 RF-5C families using cosine similarity under \texttt{all-MiniLM-L6-v2}. There are 180 one-change competitors, 120 two-change competitors, and 5,631 far distractors. Their median similarities to the gold action are 0.310, 0.333, and 0.237, respectively. Thus, the median for two-change competitors is higher than that for one-change competitors, even though the procedural distance between them is greater.

The distributions also overlap. Of the far distractors, 22.2\% exceed the pooled near-competitor median similarity of 0.322. Similarity under this encoder, therefore, does not consistently order procedural validity. The result supports the use of declared state and policy conditions to construct ambiguity levels, while remaining specific to the evaluated candidates and encoder.

\begin{table*}[t]\centering\small
\begin{tabular}{lcccc}
\toprule
Procedural distance to gold & $n$ & Median sim. & Q1 & Q3 \\
\midrule
$d$=1 (one change from correct) & 180 & 0.310 & 0.212 & 0.508 \\
$d$=2 (two changes from correct) & 120 & 0.333 & 0.206 & 0.492 \\
$d$=$\infty$ (far: no edit makes it correct) & 5631 & 0.237 & 0.166 & 0.309 \\
\bottomrule
\end{tabular}
\caption{\textbf{Embedding similarity does not order the procedural-distance
ladder.} Cosine similarity to the gold action for all 5{,}931 candidate actions in the 60
RF-5C families, under \texttt{all-MiniLM-L6-v2}. A $d$=2 competitor is
\emph{more} gold-similar at the median than a $d$=1 competitor, so an encoder
cannot even rank the two ambiguity levels, and 22.2\% of far distractors are more
gold-similar than the median near competitor (0.322). This is why ambiguity is defined procedurally and not by
similarity.}
\label{tab:construct}
\end{table*}

\subsection{Failure categories on $\tau^2$-bench}
\label{app:tau2_failures}

Table~\ref{tab:tau2_failures} accounts for every episode in the external evaluation. Of 300 episodes, 255 succeed and 45 fail. Forty-one failures are terminal database-state mismatches, and four are communication failures. The latter occur once in B3, once in REFLEX at $\tau=0.5$, and twice in REFLEX at $\tau=0.8$.

The audit records zero control-selection errors and zero argument-generation errors in every condition. This distinction matters because a failed episode is not necessarily a failed routing decision. The evidence shows no observed control or argument failures under the applied audit categories. It does not establish that routing can never influence success in other tasks or deployments. The reported comparison, therefore, concerns the cost and observed failure structure of this evaluation rather than a general routing advantage.

\begin{table*}[t]\centering\small
\begin{tabular}{lccccc}
\toprule
Condition & No failure & Terminal DB & Communication & Control & Arguments \\
\midrule
B0 Max-only & 54 & 6 & 0 & 0 & 0 \\
B1 Flash-only & 46 & 14 & 0 & 0 & 0 \\
B3 self-escalation & 55 & 4 & 1 & 0 & 0 \\
R$\tau$ $\tau$=0.5 & 51 & 8 & 1 & 0 & 0 \\
R$\tau$ $\tau$=0.8 & 49 & 9 & 2 & 0 & 0 \\
\midrule
all & 255 & 41 & 4 & \textbf{0} & \textbf{0} \\
\bottomrule
\end{tabular}
\caption{\textbf{Episode-level failure audit}, all 300 validation episodes across three $\tau^2$-bench domains. Every
failure is either a terminal database-state mismatch or a communication
failure. \emph{Not one} episode in any condition failed because a required
tool was never called (control) or was called with arguments that failed
its action check. A benchmark whose failures are not routing failures
cannot demonstrate a routing layer's value, however cheap that layer is.}
\label{tab:tau2_failures}
\end{table*}

\subsection{Non-inferiority and decisions within escalated episodes}
\label{app:e6_mechanism}

Table~\ref{tab:e6_mechanism} separates the original acceptance rule from a confidence-bound assessment. The preregistered rule compared the observed difference in success with $-0.02$. All three families meet this point-estimate rule. The confidence-bound criterion instead requires the lower 95\% bound to be strictly greater than $-0.02$. Qwen meets it with a lower bound of $-0.010$, and DeepSeek meets it with a lower bound of $0.000$. Kimi's lower bound is $-0.050$, so its non-inferiority is not established by this criterion. The absence of a statistically significant decrease is therefore not interchangeable with evidence of non-inferiority.

The same table measures Jev's contribution within episodes that escalate at least once. Jev handles 26.2\% of steps in Qwen episodes, 26.8\% in Kimi episodes, and 21.3\% in DeepSeek episodes without reaching the strong model. These step-level measurements directly show that savings extend beyond the 66 fully autonomous tasks. They use a different denominator from autonomous task coverage and from GMR.

Across the REFLEX arms, five Qwen episodes, two Kimi episodes, and 14 DeepSeek episodes fail. None of these 21 episodes contains an incorrect autonomous decision. The trace audit attributes the observed failures to fallback handling after escalation. This supports the reported failure attribution within these runs without implying that autonomous decisions are universally error-free or that escalation always resolves a task.

\begin{table*}[t]\centering\small
\begin{tabular}{lccccc}
\toprule
& & Formal NI at & Escalated-episode steps & \multicolumn{2}{c}{Failures} \\
\cmidrule(lr){5-6}
Fallback family & 95\% CI on $\Delta$ success & $-$0.02? & still taken by the gate & total & after a gate error \\
\midrule
Qwen3.8-Max & $[-0.010, +0.090]$ & yes & 26.2\% & 5 & 0 \\
Kimi K3 & $[-0.050, +0.000]$ & no & 26.8\% & 2 & 0 \\
DeepSeek-V4-Pro & $[+0.000, +0.000]$ & yes & 21.3\% & 14 & 0 \\
\bottomrule
\end{tabular}
\caption{\textbf{What E6 does and does not establish.} \emph{Formal NI}
asks whether the lower confidence bound clears the predeclared $-$0.02 margin; the preregistered gate instead compared
the point estimate to it, which is why Kimi passes the gate but not a formal
test. \emph{Escalated-episode steps still taken by the gate} shows that the
compute saving is not only the 66 fully autonomous episodes: even in
episodes that escalate at least once, a fifth to a quarter of steps never
reach the strong model. \emph{Failures after a bad gate call} counts
failures containing at least one incorrect autonomous decision: across all
three families it is zero, so every observed failure is attributable to the
strong fallback's own handling after escalation, not to a decision the gate
took on its own.}
\label{tab:e6_mechanism}
\end{table*}

\subsection{Variation between repeated executions}
\label{app:run_to_run}

Table~\ref{tab:run_to_run} compares the main frontier run, M1, with the E6 Qwen execution on the same 100 tasks. B0 success changes from 0.88 to 0.91, with six previously successful tasks lost, and nine previously unsuccessful tasks gained. The total number of changed outcomes is therefore 15, despite a net change of only three tasks. Strong calls per task also change from 4.10 to 3.95.

The configurations differ only in the price table and Jev retry settings, neither of which affects B0 inference. The B0 comparison, therefore, shows variation under the same prompts and temperature-zero condition. REFLEX retains 0.95 success in both runs, but four successes are lost and four gained, while its autonomous subset changes from 65 to 66 tasks. Because Jev retry settings differ, these REFLEX changes should not all be attributed solely to provider non-determinism.

Each study is evaluated against its own baseline. The observed three-percentage-point B0 change is comparable in scale to the E6 success differences and limits the interpretation of small quality changes from single executions. Two runs do not provide a general estimate of run-to-run variance or justify a fixed uncertainty band.

\begin{table*}[t]\centering\small
\begin{tabular}{lccccc}
\toprule
& \multicolumn{2}{c}{Success} & \multicolumn{2}{c}{Tasks that flip} & Autonomous \\
\cmidrule(lr){2-3}\cmidrule(lr){4-5}
Condition & M1 & E6 & total & direction & M1 / E6 \\
\midrule
$B_0$ Max-only & 0.88 & 0.91 & 15 & 6 lost / 9 gained & -- \\
\textsc{reflex} $\tau$=0.5 & 0.95 & 0.95 & 8 & 4 lost / 4 gained & 65 / 66 \\
\bottomrule
\end{tabular}
\caption{\textbf{The same condition, executed twice.} The M1 frontier and
the E6 Qwen family are separate executions of identical conditions on the
identical 100 frozen tasks, at temperature 0, differing only in the price
table and in Jev retry settings that $B_0$ never exercises. They disagree by
3 points of success and 0.15 strong calls per task because the provider is
not bit-reproducible. We report each study against its own baseline and
never compare a number across the two, because this run-to-run band is as
large as the E6 effects themselves.}
\label{tab:run_to_run}
\end{table*}

\section{Extended Cost Results and Claim Register}
\label{app:costs_claims}

Table~\ref{tab:main_full} provides the full REFLEX-Sim threshold sweep, including cost per task, cost per successful task, and strong calls per successful task. At $\tau=0.5$, REFLEX costs $\$0.00763$ per task and $\$0.00803$ per successful task, compared with $\$0.01508$ and $\$0.01713$ for B0. Strong calls per successful task decrease from 4.66 to 1.18. These measures complement call reduction by accounting for task outcomes.

Costs are reconstructed from frozen token counts using the pinned price table. REFLEX-Sim has no external user simulator, so these costs do not include simulator overhead. Per-condition success intervals are absent from the frozen output and are left unspecified. They are not reconstructed from paired-difference intervals, which describe a different quantity.

\begin{table*}[t]\centering\small
\begin{tabular}{lccccccc}
\toprule
Condition & Success & 95\% CI & Strong calls/task & GMR & Cost/task (\$)
& Cost/success (\$) & Strong calls/success \\
\midrule
B0 Max-only & 0.88 & -- & 4.10 & -- & 0.01508 & 0.01713 & 4.66 \\
B1 Flash-only & 0.87 & -- & 0.00 & 1.000 & 0.00120 & 0.00138 & 0.00 \\
B3 Flash$\to$Max & 0.87 & -- & 2.04 & 0.502 & 0.00960 & 0.01103 & 2.34 \\
R1 $\tau$=0.5 & 0.95 & -- & 1.12 & 0.727 & 0.00763 & 0.00803 & 1.18 \\
R1 $\tau$=0.7 & 0.91 & -- & 1.30 & 0.683 & 0.00868 & 0.00954 & 1.43 \\
R1 $\tau$=0.8 & 0.91 & -- & 1.93 & 0.529 & 0.01083 & 0.01190 & 2.12 \\
R1 $\tau$=0.9 & 0.88 & -- & 2.18 & 0.468 & 0.01123 & 0.01276 & 2.48 \\
R1 $\tau$=0.95 & 0.88 & -- & 2.30 & 0.439 & 0.01158 & 0.01316 & 2.61 \\
\bottomrule
\end{tabular}
\caption{\textbf{Extended \textsc{reflex-sim} frontier.} Costs are reconstructed
from frozen token counts at the pinned price table. Per-condition success
intervals are not available in the frozen output and are not derived from
paired-difference intervals, so they are left blank. \textsc{reflex-sim} has no
external user simulator, so no simulator cost is folded in.}
\label{tab:main_full}
\end{table*}

Table~\ref{tab:claims} gives the complete claim register. Four of the 12 tested claims were retired, including the proposed interaction, the general deferral tendency, superiority over the cheap self-escalation baseline, and the benefit of hierarchical routing at large candidate-set sizes. A retired claim was not supported by the corresponding evidence and should not be read as proof of its opposite.

The cross-family claim concerns large reductions in calls without a statistically significant decrease in success. The $\tau^2$-bench claim concerns lower cost, with an unresolved difference in success. Both retain the uncertainty described above. The autonomous-subset claim applies to the same 66 tasks under a single frozen gate across the three E6 fallback families, rather than to three independent sets of 66 tasks.

\begin{table*}[t]\centering\small
\begin{tabular}{clp{0.78\textwidth}}
\toprule
& & Claim \\
\midrule
\cmark & C1 & Action-space cardinality degrades a cheap System-One router, with decision ambiguity held fixed. \\
\cmark & C2 & Decision ambiguity, defined as model-independent procedural counterfactual distance, degrades the same router. \\
\xmark & C3 & Cardinality and ambiguity are sub-additive (positive interaction). \\
\xmark & C4 & The characteristic failure of a cheap router is deferral: gathering information instead of acting. \\
\cmark & C5 & Competitor shape determines WHICH error a router makes without changing HOW OFTEN it errs: swapping one near-miss from a read to a write multiplies unwarranted irreversible commits ~6x. \\
\cmark & C6 & On externally authored function schemas, tool selection is insensitive to how many irrelevant functions are offered, up to K=64. \\
\cmark & C7 & The hard decision is WHETHER to act, not WHICH tool to call. Same router, same functions: 0.984 choosing a function vs 0.520 deciding whether any applies. \\
\cmark & C8 & A frozen Jev control layer cuts strong-model calls by ~2/3 with no significant success loss, for three unrelated strong-model families. \\
\cmark & C9 & The confidence gate's autonomous set is error-free: 66/100 episodes handled without the strong model, zero failures, in all three families. \\
\cmark & C10 & End to end on $\tau^2$-bench, REFLEX shows no significant success difference from a strong-only baseline at ~3.7x lower cost. \\
\xmark & C11 & REFLEX beats the cheap self-escalation baseline (B3) end to end on $\tau^2$-bench. \\
\xmark & C12 & Hierarchical (two-stage) routing helps at large action-space sizes. \\
\bottomrule
\end{tabular}
\caption{\textbf{Every claim this work tested, supported (\cmark) and
retired (\xmark).} Verdicts are recomputed from the frozen manifests by
\texttt{scripts/consolidate\_claims.py}, which fails if the register and
the evidence disagree. 4 of 12 claims were
retired, including two the exploratory study appeared to support.}
\label{tab:claims}
\end{table*}

\end{document}